\documentclass[runningheads]{llncs}
\usepackage{graphicx}
\usepackage{amsmath,amssymb} 
\usepackage{color}
\usepackage[width=122mm,left=12mm,paperwidth=146mm,height=193mm,top=12mm,paperheight=217mm]{geometry}

\usepackage{multicol}
\usepackage{mathtools}
\usepackage[table]{xcolor}

\DeclareMathOperator*{\argmax}{arg\,max}

\newcolumntype{L}[1]{>{\raggedright\let\newline\\\arraybackslash\hspace{0pt}}m{#1}}
\newcolumntype{C}[1]{>{\centering\let\newline\\\arraybackslash\hspace{0pt}}m{#1}}
\newcolumntype{R}[1]{>{\raggedleft\let\newline\\\arraybackslash\hspace{0pt}}m{#1}}

\begin{document}
\pagestyle{headings}
\mainmatter

\title{Modeling Context Between Objects for Referring Expression Understanding}

\titlerunning{Modeling Context Between Objects for Referring Expression Understanding}

\authorrunning{Varun K. Nagaraja, Vlad I. Morariu, Larry S. Davis}
\author{Varun K. Nagaraja \quad Vlad I. Morariu \quad Larry S. Davis}
\institute{University of Maryland, College Park, MD, USA. \\ 
\small \texttt{\{varun,morariu,lsd\}@umiacs.umd.edu}}

\maketitle

\begin{abstract}
Referring expressions usually describe an object using properties of the object and relationships of the object with other objects. We propose a technique that integrates context between objects to understand referring expressions. Our approach uses an LSTM to learn the probability of a referring expression, with input features from a region and a context region. The context regions are discovered using multiple-instance learning (MIL) since annotations for context objects are generally not available for training. We utilize max-margin based MIL objective functions for training the LSTM. Experiments on the Google RefExp and UNC RefExp datasets show that modeling context between objects provides better performance than modeling only object properties. We also qualitatively show that our technique can ground a referring expression to its referred region along with the supporting context region.
\end{abstract}

\section{Introduction}
In image retrieval and human-robot interaction, objects are usually queried by their category, attributes, pose, action and their context in the scene \cite{Johnson_2015_CVPR}.
Natural language queries can encode rich information like relationships that distinguish object instances from each other.
In a retrieval task that focuses on a particular object in an image, the query is called a \textit{referring expression} \cite{ReferItGame,Krahmer12}.
When there is only one instance of an object type in an image, a referring expression provides additional information such as attributes to improve retrieval/localization performance.
More importantly, when multiple instances of an object type are present in an image, a referring expression distinguishes the referred object from other instances, thereby helping to localize the correct instance.
The task of localizing a region in an image given a referring expression is called the \textit{comprehension task} \cite{Mao15} and its inverse process is the \textit{generation task}. 
In this work we focus on the comprehension task.

\begin{figure}[t]
\centering
\includegraphics[width=0.975\textwidth]{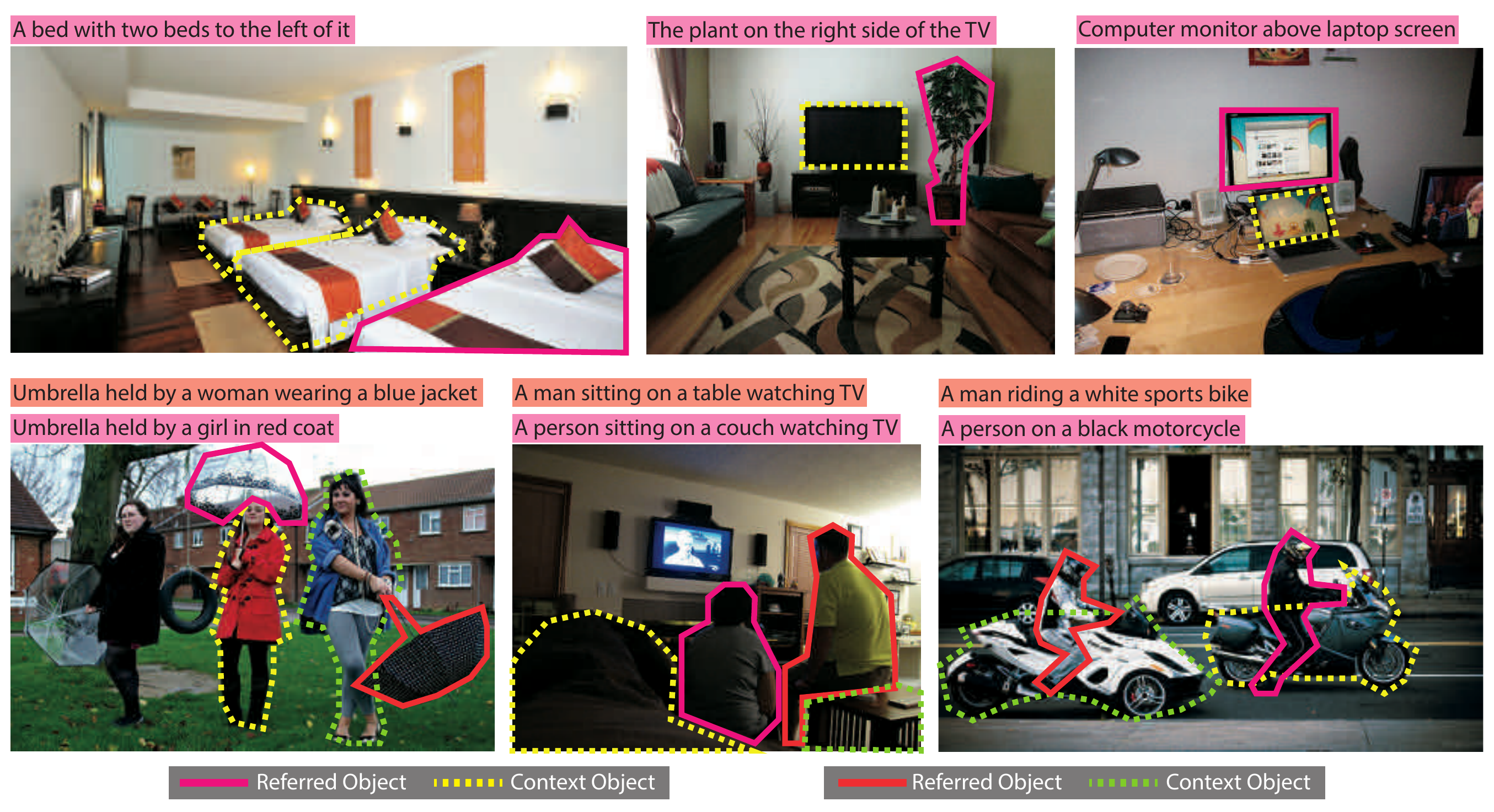}
\caption{
Context between objects is specified using spatial relationships between regions such as ``above", ``to the right", ``to the left" etc.
It is also represented using interactions between objects such as ``riding", ``holding" etc.
When there are multiple instances of the same type of object, context helps in referring to the appropriate instance.
}
\label{Fig:MotivationImages}
\end{figure}

Referring expressions usually mention relationships of an object with other regions along with the properties of the object \cite{Mitchell2010,Viethen2008} (See Figure \ref{Fig:MotivationImages}).
Hence, it is important to model relationships between regions for understanding referring expressions.
However, the supervision during training typically consists of annotations of only the referred object.
While this might be sufficient for modeling attributes of an object mentioned in a referring expression, it is difficult to model relationships between objects with such limited supervision.
Previous work on referring expressions \cite{ReferItGame,Mao15,HuXRFSD15} generally ignores modeling relationships between regions.
In contrast, we learn to map a referring expression to a region and its supporting context region.
Since the bounding box annotations of context objects are not available for training, we learn the relationships in a weakly supervised framework.

We follow the approach of Mao et al. \cite{Mao15} to perform the comprehension task. 
The probability of a referring expression is measured for different region proposals and the top scoring region is selected as the referred region. 
The input features in our model are obtained from a \textit{\{region, context\_region\}} pair where the image itself is considered as one of the context regions.
The probability of a referring expression for a region can then be pooled over multiple pairs using the max function or the noisy-or function. 
We use an LSTM \cite{LSTM} for learning probabilities of a referring expression similar to Mao et al. \cite{Mao15}.
Since the bounding boxes for context objects are not known during training, we train using a Multiple-Instance Learning (MIL) objective function.
The max-margin based LSTM training of Mao et al. \cite{Mao15} is extended to max-margin MIL training for LSTMs.
The first formulation is similar to MI-SVM \cite{MISVM} which has only negative bag margin and the second formulation is similar to mi-SVM \cite{MISVM} which has both positive and negative bag margins.
Experiments are performed on the Google~RefExp dataset \cite{Mao15} and UNC RefExp dataset \cite{UNCRefExp}.
Our results show that modeling objects in context for the comprehension task provides better performance than modeling only object properties.
We also qualitatively show that our technique can ground the correct context regions for those referring expressions which mention object relationships.


\section{Related Work}
The two tasks of localizing an object given a referring expression and generating a referring expression given an object are closely related. 
Some image caption generation techniques \cite{Vinyals_2015_CVPR,plummer2015flickr30k} first learn to ground sentence fragments to image regions and then use the learned association to generate sentences. 
Since the caption datasets (Flickr30k-original \cite{flickr30k}, MS-COCO \cite{MSCOCO}) do not contain the mapping from phrases to object bounding boxes, the visual grounding is learned in a weakly supervised manner.
Fang et al. \cite{Fang15} use multiple-instance learning to learn the probability of a region corresponding to different words.
However, the associations are learned for individual words and not in context with other words.
Karpathy et al. \cite{Karpathy_2015_CVPR} learn a common embedding space for image and sentence with an MIL objective such that a sentence fragment has a high similarity with a single image region. 
Instead of associating each word to its best region, they use an MRF to encourage neighboring words to associate to common regions.

Attention based models implicitly learn to select or weigh different regions in an image based on the words generated in a caption. 
Xu et al. \cite{XuBKCCSZB15} propose two types of attention models for caption generation.
In their stochastic hard attention model, the attention locations vary for each word and in the deterministic soft attention model, a soft weight is learned for different regions. 
Neither of these models are well suited for localizing a single region for a referring expression.
Rohrbach et al. \cite{RohrbachRHDS15} learn to ground phrases in sentences using a two stage model. 
In the first stage, an attention model selects an image region and in the second stage, the selected region is trained to predict the original phrase.
They evaluate their technique on the Flickr 30k Entities dataset \cite{plummer2015flickr30k} which contains mappings for noun phrases in a sentence to bounding boxes in the corresponding image. 
The descriptions in this dataset do not always mention a salient object in the image.
Many times the descriptions mention groups of objects and the scene at a higher level and hence it becomes challenging to learn object relationships.

Kong et al. \cite{KongLBUF14} learn visual grounding for nouns in descriptions of indoor scenes in a supervised manner. They use an MRF which jointly models scene classification, object detection and grounding to 3D cuboids. Johnson et al. \cite{JohnsonKL15} propose an end-to-end neural network that can localize regions in an image and generate descriptions for those regions. 
Their model is trained with full supervision with region descriptions present in the Visual Genome dataset \cite{VisGenome}. 

Most of the works on referring expressions learn to ground a single region by modeling object properties and image level context. 
Rule based approaches to generating referring expressions \cite{mitchell2011two,fitzgerald2013learning} are restricted in the types of properties that can be modeled.
Kazemzadeh et al. \cite{ReferItGame} designed an energy optimization model for generating referring expressions in the form of object attributes.
Hu et al. \cite{HuXRFSD15} propose an approach with three LSTMs which take in different feature inputs such as region features, image features and word embedding.
Mao et al. \cite{Mao15} propose an LSTM based technique that can perform both tasks of referring expression generation and referring expression comprehension.
They use a max-margin based training method for the LSTM wherein the probability of a referring expression is high only for the referred region and low for every other region.
This type of training significantly improves performance.
We extend their max-margin approach to multiple-instance learning based training objectives for the LSTM.
Unlike previous work, we model context between objects for comprehending referring expressions.

\section{Modeling context between objects}
Given a referring expression $S$ and an image $I$, the goal of the comprehension task is to predict the (bounding box of the) region $R^*$ that is being referred to. 
We adopt the method of Mao et al. \cite{Mao15} and start with a set of region proposals ($\mathcal{C}$) from the image.
We learn a model that measures the probability of a region given a referring expression.
The maximum scoring region $R^* = \argmax_{R \in \mathcal{C}} p(R | S, I)$ is then selected as the referred region.
Mao et al. \cite{Mao15} rewrite the scoring function as $R^* = \argmax_{R \in \mathcal{C}} p(S | R, I)$ by applying Bayes' rule and assuming a uniform prior for $p(R|I)$.
This implies that comprehension can be accomplished using a model trained to generate sentences for an image region.

Many image and video captioning techniques \cite{Vinyals_2015_CVPR,LRCN,Venugopalan_2015_ICCV}, learn the probability of a sentence given an image or video frame using an LSTM. 
The input features to the LSTM consist of a word embedding vector and CNN features extracted from the image. 
The LSTM is trained to maximize the likelihood of observing the words of the caption corresponding to the image or the region. 
This model is used by Mao et al. \cite{Mao15} as the baseline for referring expression comprehension. 
Along with the word embedding and region features, they also input CNN features of the entire image and bounding box features to act as context. 
They further propose a max-margin training method for the LSTM to enforce the probability of a referring expression to be high for the referred region and low for all other regions. 
For a referring expression $S$, let $R_n \in \mathcal{C}$ be the true region and $R_i \in \mathcal{C}\setminus R_n$ be a negative region; then the training loss function with a max-margin component is written as
\begin{align}
J(\theta) = - \sum_{R_i \in \mathcal{C}\setminus R_n} 
\left\{
\begingroup
\renewcommand*{\arraystretch}{1.25}
\begin{matrix*}[l]
\log p(S | R_n,I,\theta)\\
- \lambda \max(0, M-\log p(S | R_n,I,\theta) + \log p(S | R_i,I,\theta)
\end{matrix*}
\endgroup
\right\}
\label{Eqn:MaxMargin}
\end{align}
where $\theta$ are the parameters of the model, $\lambda$ is the weight for the margin loss component and $M$ is the margin. The max-margin model has the same architecture as the baseline model but is trained with a different loss function.

\begin{figure}[t]
\centering
\includegraphics[width=\textwidth]{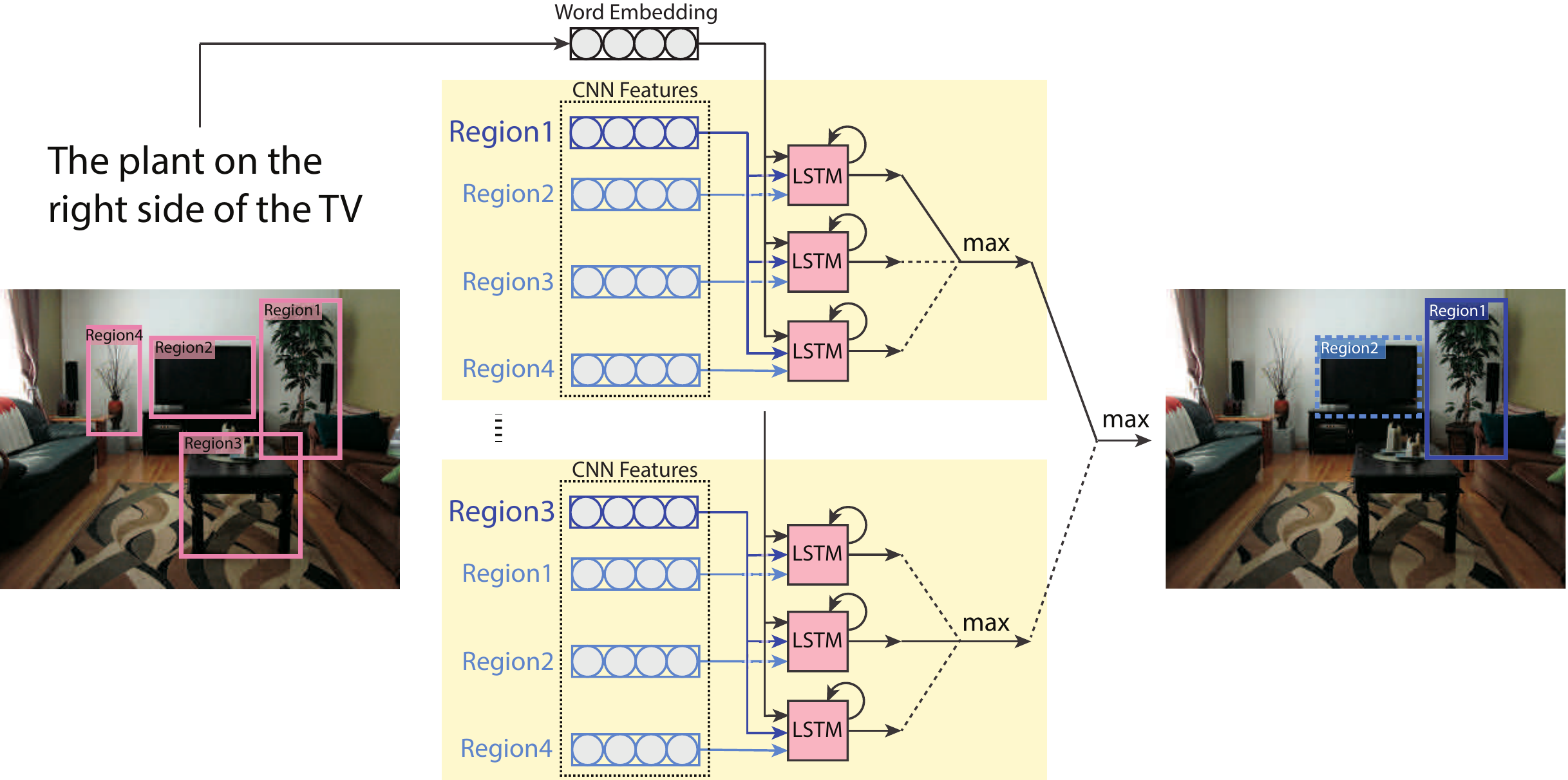}
\caption{We identify the referred region along with its supporting context region. We start with a set of region proposals in an image and consider pairs of the form \textit{\{region, context\_region\}}. The entire image is also considered as a potential context region. The probability is evaluated using an LSTM which takes as input region CNN features, context region CNN features, bounding box features and an embedding vector for words in the referring expression. All the LSTMs share the same weights. The probability of a referring expression for an individual region is obtained by finding the maximum over its pairs with context regions. The noisy-or function can be used instead of the max function. After pooling over context regions, the top scoring region (along with its context region) is selected as the referred region}
\label{Fig:Overview}
\end{figure}

In the above model, the probability of a referring expression is influenced by the region and only the image as context. 
However, many referring expressions mention an object in relation to some other object (e.g., ``The person next to the table") and hence it is important to incorporate context information from other regions as well. 
One of the challenges for learning relationships between regions through referring expressions is that the annotations for the context regions are generally not available for training.
However, we can treat combinations of regions in an image as bags and use Multiple Instance Learning (MIL) to learn the probability of referring expressions.
MIL has been used by image captioning techniques \cite{Fang15,Karpathy_2015_CVPR,karpathy2014} to associate phrases to image regions when the ground-truth mapping is not available.

We learn to map a referring expression to a region and its supporting context region.
We start with a set of region proposals in an image and consider pairs of the form \textit{\{region, context\_region\}}. 
The image is included as one of the context regions.
The probability of a referring expression is learned for pairs of regions where the input features include visual features and bounding box features for both regions.
The probability of an individual region is then obtained by pooling from probabilities of the region's combinations with its potential context regions. 
After pooling, the top scoring region (along with its context region) is selected as the referred region.
Figure \ref{Fig:Overview} shows an overview of our system.

Let $\mathcal{C} = \{I,R_1,R_2,\dots,R_n\}$ be the set of candidate context regions which includes the entire image, $I$, and other regions generated by the object proposal algorithm.
The minimum size of the context region set is one since it always includes $I$ and the model in that case would be equivalent to Mao et al. \cite{Mao15}.
We now define the probability of a sentence $S$ given a region $R$ as 
\begin{equation}
p(S|R) = \max_{R_i \in \mathcal{C}\setminus R} p(S|R,R_i)
\label{Eqn:MaxPooledProb}
\end{equation}
This implies that the probability of a sentence given a region is defined as the maximum probability obtained by any of the region's combination with a context region. 
The referred region can now be selected as the top scoring region from the max-pooled probabilities.
\begin{equation}
R^* = \argmax_{R \in \mathcal{C}\setminus I} \left\{ \max_{R_i \in \mathcal{C}\setminus R} p(S|R,R_i) \right\}
\label{Eqn:MILMaxSelector}
\end{equation}
The noisy-or function can be used instead of the max function in Equation \ref{Eqn:MaxPooledProb}. 
Then the referred region is selected as
\begin{equation}
R^* = \argmax_{R \in \mathcal{C}\setminus I} \left\{ 1 - \prod_{R_i \in \mathcal{C}\setminus R} (1 - p(S|R,R_i)) \right\}
\label{Eqn:MILNoisyOrSelector}
\end{equation}
The noisy-or function can integrate context information from more than one pair of regions and it is more robust to noise than the max function.

\begin{figure}[t]
\centering
\includegraphics[width=\textwidth]{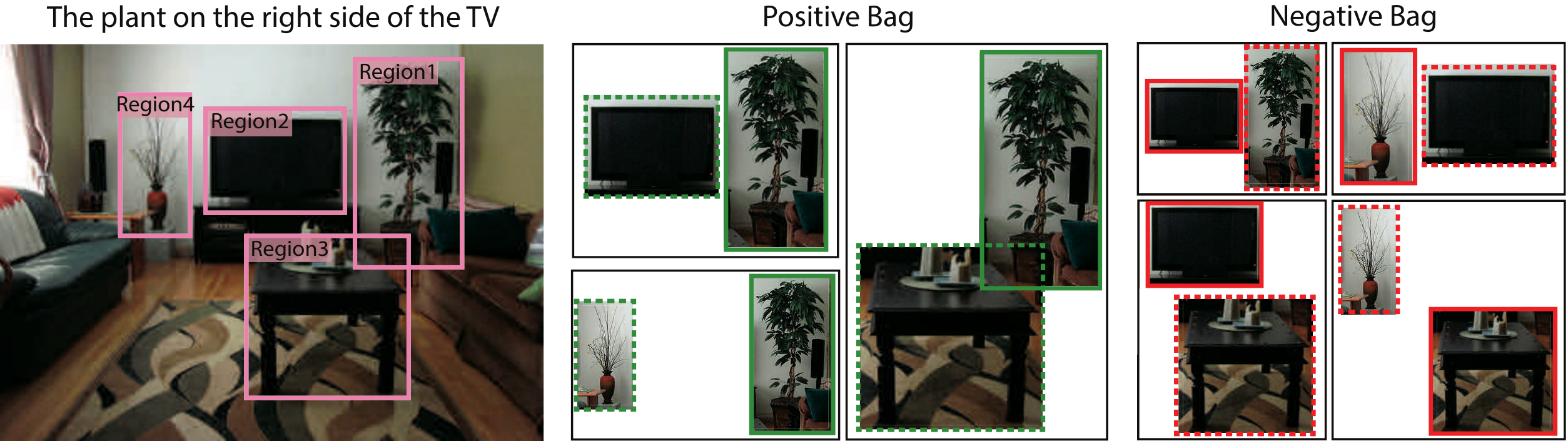}
\caption{Given a set of region proposals in an image, we construct positive and negative bags containing pairs of regions.
In this example, the plant in $Region1$ is the referred object.
Hence the positive bag consists of pairs of the form ($Region1$,$R_i$) where $R_i$ is one of the remaining regions.
The negative bag consists of pairs of the form $(R_i, R_j)$ where the first region $R_i$ can be any region except $Region1$ and the second region $R_j$ can be any region including $Region1$}
\label{Fig:Bags}
\end{figure}

We learn the probability function $p(S|R_i,R_j)$ using multiple-instance learning.
In our MIL framework, a positive bag for a referring expression consists of pairs of regions of the form $(R_t,R_i)$.
The first element in the pair is the region $R_t$ referred to in the expression and the second element is a context region $R_i \in \mathcal{C}\setminus R_t$.
A negative bag consists of pairs of regions of the form $(R_i,R_j)$ where $R_i \in \mathcal{C}\setminus R_t$ and $R_j \in \mathcal{C}$.
Figure \ref{Fig:Bags} shows an example of bags constructed for a sample referring expression.
 
An LSTM is used to learn the probability of referring expressions and we define multiple-instance learning objective functions for training.
Similar to the max-margin training objective defined in Equation \ref{Eqn:MaxMargin}, we apply the max-margin approach of MI-SVM and mi-SVM \cite{MISVM} here to train the LSTM. 
In MI-SVM, the margin constraint is enforced on all the samples from the negative bag but only on the positive instances from the positive bag. 
The training loss function with a margin for the negative bag is given by 
\begin{align}
J'(\theta) = - \sum_{\substack{R_i \in \mathcal{C}\setminus R_t, \\ R_j \in \mathcal{C}}}
\left\{
\begingroup
\renewcommand*{\arraystretch}{1.25}
\begin{matrix*}[l]
\log p(S | R_t,\theta)\\ 
- \lambda_N \max(0, M- \log p(S | R_t,\theta) + \log p(S | R_i,R_j,\theta)
\end{matrix*}
\endgroup
\right\}
\label{Eqn:NegBagMargin}
\end{align}
The difference between the max-margin Equation~\ref{Eqn:MaxMargin} and Equation~\ref{Eqn:NegBagMargin} is that the probability of the referred region is now obtained from Equation~\ref{Eqn:MaxPooledProb} and the negative samples are not just pairs of regions with the entire image.

The loss function in Equation \ref{Eqn:NegBagMargin} ignores potential negative instances in the positive bag.
We can attempt to identify the negative instances and apply a margin to those pairs as well.
In mi-SVM, the labels for instances in positive bags are assumed to be latent variables.
The goal is to maximize the margin between all positive and negative instances jointly over the latent labels and the discriminant hyperplane.
In many referring expressions, there is usually one other object mentioned in context. 
We assume that there is only one positive pair in the positive bag and assign a positive label for the instance with the maximum probability.
The remaining pairs in the positive bag are assigned a negative label.
Without loss of generality, let $(R_t, R_c)$ be the positive instance from the positive bag.
The training loss function with margins for both positive and negative bags is given by,
\begin{align}
J''(\theta) &= - \sum_{\substack{R_i \in \mathcal{C}\setminus R_t, \\ R_j \in \mathcal{C}}}
\left\{
\begingroup
\renewcommand*{\arraystretch}{1.25}
\begin{matrix*}[l]
\log p(S | R_t, R_c,\theta) \\ 
- \lambda_N \max(0, M- \log p(S | R_t,R_c,\theta) + \log p(S | R_i,R_j,\theta)
\end{matrix*}
\endgroup
\right\} \nonumber \\
& \quad - \sum_{\substack{R_k \in \mathcal{C} \setminus R_c}}
\left\{
\begingroup
\renewcommand*{\arraystretch}{1.25}
\begin{matrix*}[l]
\log p(S | R_t, R_c,\theta) \\ 
- \lambda_P \max(0, M- \log p(S | R_t,R_c,\theta) + \log p(S | R_t,R_k,\theta)
\end{matrix*}
\endgroup
\right\}
\label{Eqn:PosNegBagMargin}
\end{align}
In the training algorithm proposed by Andrews et al. \cite{MISVM} for mi-SVM, the latent labels for instances in a positive bag are obtained in an iterative manner.
The mi-SVM algorithm iterates over two steps: use the current hyperplane to determine the latent labels, then use the labels to train a new hyperplane. 
Since neural networks are trained over multiple epochs of the data, the training process is similar to the iterative algorithm used to train mi-SVM.
During an epoch, the positive instance $(R_t, R_c)$ in the positive bag is determined as 
\begin{equation}
R_c = \argmax_{R_i \in \mathcal{C}\setminus R_t} p(S|R_t,R_i)
\end{equation}
The parameter $\theta$ is updated by applying the loss function in Equation \ref{Eqn:PosNegBagMargin} with $R_c$ substituted into it. 
In the following epoch, $R_c$ is updated using the model with updated parameter $\theta$.

The assumption that there is one positive instance in the positive bag holds true when a referring expression uniquely identifies an object and its context object.
Such referring expressions are present in the Google RefExp dataset (e.g., ``A white truck in front of a yellow truck").
The UNC RefExp dataset contains referring expressions which do not always uniquely refer to an object with its context object (e.g., ``Elephant towards the back").
Hence the two different formulations (Equation \ref{Eqn:NegBagMargin} and Equation \ref{Eqn:PosNegBagMargin}) harness different characteristics of referring expressions between the two datasets.

\section{Experiments}

\subsection{Datasets}
We perform experiments on the Google RefExp dataset \cite{Mao15} and the UNC RefExp dataset \cite{UNCRefExp}.
Both datasets contain referring expressions for images in the Microsoft COCO dataset \cite{MSCOCO}.

The dataset partition accompanying the current release of Google RefExp dataset was created by randomly selecting 5000 objects for validation and 5000 objects for testing. 
This type of partitioning results in overlapping images between training, validation and test sets. 
To avoid any overlap between the partitions, we create our own partition for the training and validation sets. 
Our training partition contains 23199 images with 67996 objects. 
Some objects have multiple referring expressions and hence the total number of referring expressions is 85,408.
The validation partition contains 2600 images with 7623 objects and 9602 referring expressions.
The results of the baseline and max-margin techniques did not differ much between our partition and the Mao et al. \cite{Mao15} partition.
However, we perform experiments with our partition since we model context from many regions in an image and that information should not leak into the test stage.
We will make our partition publicly available. 
The test set of this dataset has not been released yet. 
Hence, we use 4800 referring expressions from the training set for validation.

The UNC RefExp dataset was collected by applying the ReferIt game \cite{ReferItGame} on MS-COCO images. 
The training partition contains 16994 images, 42404 objects and 120624 referring expressions.
The validation partition contains 1500 images, 3811 objects and 10834 referring expressions.
The testing partition contains two splits.
TestA partition contains 750 images, 1975 objects and 5657 person-centric referring expressions.
TestB partition contains 750 images, 1810 objects and 5095 object-centric referring expressions.
While Mao et al. \cite{Mao15} create their own test partition of the UNC RefExp data from a random subset of objects, we work with the partitioning provided by Yu et al. \cite{UNCRefExp}.

The evaluation is performed by measuring the Intersection over Union (IoU) ratio between a groundtruth box and the top predicted box for a referring expression. If the IoU \textgreater 0.5, the prediction is considered a true positive and this is the Precision@1 score. The scores are then averaged over all referring expressions.

\subsection {Implementation details}
Our neural network architecture is the same as Mao et al. \cite{Mao15}.
We use an LSTM to learn probabilities of referring expressions. 
The size of the hidden state vector is 1024.
We extract CNN features for a region and its context region using the 16 layer VGGNet \cite{VGGNet} pre-trained on the ImageNet dataset. 
We use the 1000 dimensional features from the last layer (fc8) of VGGNet and fine tune only the last layer while keeping everything else fixed.
The CNN features for each region are concatenated with bounding box features of the form $[\frac{x_{min}}{W},\frac{y_{min}}{H},\frac{x_{max}}{W},\frac{y_{max}}{H},\frac{Area_{bbox}}{Area_{image}}]$ where $(W,H)$ are the width and height of the image.
The resulting feature length for both the region and the context region is 2010.
We scale the features to lie between -0.5 and 0.5 before feeding them into the LSTM.
The scaling factors were obtained from the training set.
We use a vector embedding of size 1024 for the words in a referring expression.
The size of the vocabulary is 3489 and 2020 for the Google RefExp and UNC RefExp datasets respectively. 
The vocabularies are constructed by choosing words that occur at least five times in the training sets.
We also filter out special characters of length 1.

We implement our system using the Caffe framework \cite{Caffe} with LSTM layer provided by Donahue et al. \cite{LRCN}.
We train our network using stochastic gradient descent with a learning rate of 0.01 which is halved every 50,000 iterations. 
We use a batch size of 16.
The word embedding and LSTM layer outputs are regularized using dropout with a ratio of 0.5. 

While Mao et al. \cite{Mao15} used proposals from the Multibox \cite{Multibox} technique, we use proposals from the MCG \cite{MCG} technique.
We obtain top 100 proposals for an image using MCG and evaluate scores for the 80 categories in the MS-COCO \cite{MSCOCO} dataset.
We then discard boxes with low values.
The category scores are obtained using the 16 layers VGGNet \cite{VGGNet} CNN fine-tuned using Fast RCNN \cite{fastrcnn}.
The category scores of proposals are not used during the testing stage by the referring expression model.

\subsection{Comparison of different techniques}
We compare our MIL based techniques with the baseline and max-margin models of Mao et al \cite{Mao15}.
The model architecture is the same for all the different variants of training objective functions.

Our implementation of the max-margin technique provided better results than those reported in Mao et al. \cite{Mao15}. 
We use a margin $M=0.1$ and margin weight $\lambda=1$ in the max-margin loss function. 
The margin is applied on word probabilities in the implementation.
For each referring expression and its referred region, we sample 5 ``hard MCG negatives" for training, similar to their ``hard Multibox negatives".
The ``hard MCG negatives" are MCG proposals that have the same predicted object category as the referred region. 
The object category of a proposal is obtained during the proposal filtering process. 
For our MIL based loss functions, we randomly sample 5 ground-truth proposals as context regions for training.
We also sample 5 hard MCG negatives.
We use a margin $M=0.1$ and margin weights $\lambda_N=1, \lambda_P=1$ in the MIL based loss functions.
During testing, we combine the scores from different context regions using the noisy-or function (Equation \ref{Eqn:MILNoisyOrSelector}).
We sample a maximum of 10 regions for context during the testing stage.

\begin{table*}[t]
\centering
\caption{Precision@1 score of different techniques. The results are obtained using the noisy-or function for pooling context information from multiple pairs. We experiment with both ground-truth (GT) and MCG proposals
}
\begin{multicols}{2}
\begin{center}
\begin{tabular}{|l|c|c|}
\hline
Proposals           &  GT  & MCG  \\ \hline
\multicolumn{3}{|c|}{\cellcolor[gray]{0.8} Google RefExp - Val}              \\ \hline
Max Likelihood \cite{Mao15}          & 57.5 & 42.4 \\ \hline
Max-Margin \cite{Mao15}         & 65.7 & 47.8 \\ \hline
Ours, Neg.Bag Margin       & \bf{68.4} & 49.5 \\ \hline
Ours, Pos. \& Neg. Bag Mgn.    & \bf{68.4} & \bf{50.0} \\ \hline
\multicolumn{3}{|c|}{\cellcolor[gray]{0.8} UNC RefExp - Val}                \\ \hline
Max Likelihood \cite{Mao15}           &67.5 & 51.8 \\ \hline
Max-Margin \cite{Mao15}          & 74.4 & 56.1 \\ \hline
Ours, Neg. Bag Margin      & \bf{76.9} & {57.3} \\ \hline
Ours, Pos. \& Neg. Bag Mgn.   & 76.1 & \bf{57.4} \\ \hline
\end{tabular}
\end{center}
\columnbreak
\begin{center}
\begin{tabular}{|l|c|c|}
\hline
Proposals           &  GT  & MCG  \\ \hline
\multicolumn{3}{|c|}{\cellcolor[gray]{0.8} UNC RefExp - TestA}                \\ \hline
Max Likelihood \cite{Mao15}           &65.9 & 53.2 \\ \hline
Max-Margin \cite{Mao15}          & 74.9 & 58.4 \\ \hline
Ours, Neg. Bag Margin      & \bf{75.6} & {58.6} \\ \hline
Ours, Pos. \& Neg. Bag Mgn.   & 75.0 & \bf{58.7} \\ \hline
\multicolumn{3}{|c|}{\cellcolor[gray]{0.8} UNC RefExp -TestB}               \\ \hline
Max Likelihood \cite{Mao15}           & 70.6 & 50.0 \\ \hline
Max-Margin \cite{Mao15}         & 76.3 &55.1 \\ \hline
Ours, Neg. Bag Margin       & \bf{78.0} & \bf{56.4} \\ \hline
Ours, Pos. \& Neg. Bag Mgn.   & 76.1 & 56.3 \\ \hline
\end{tabular}
\end{center}
\end{multicols}
\label{Tab:SOTAComparison}
\end{table*}

Table \ref{Tab:SOTAComparison} shows the Precision@1 scores for the different partitions of both datasets.
We show results using ground-truth proposals and MCG proposals to observe the behavior of our framework with and without proposal false positives.
The results show that our MIL loss functions perform significantly better than the max-margin technique of Mao et al. \cite{Mao15} on the validation partitions of both datasets and the TestB partition of UNC RefExp dataset.
The results on the TestA partition show only a small improvement over the max-margin technique and we investigate this further in the ablation experiments.

We observe on the Google RefExp dataset that the MIL loss function with both positive and negative bag margin performs better than the one with negative bag margin only.
In this dataset, referring expressions which mention context between objects usually identify an object and its context object uniquely.
Hence there is only one positive instance in the positive bag of region and context region pairs.
This property of the referring expressions satisfies the assumption for using the loss function with both positive and negative bag margin.

On the UNC RefExp dataset, we observe that the MIL loss function with negative bag margin performs better or similar to the loss function with both positive and negative bag margin.
Unlike the Google RefExp dataset, the referring expressions in the dataset do not always uniquely identify a context object. 
Many times the context object is not explicitly mentioned in a referring expression
e.g., in Figure \ref{Fig:Berg_pos_results}b, the elephant in the front is implied to be context but not explicitly mentioned.
The assumption of one positive instance in the positive bag does not always hold.
Hence, the performance is better using the loss function with negative bag margin only.

\begin{table}[t]
\centering
\caption{Pooling context in different ways during testing. We compare the performance of pooling context using noisy-or function, max function and also restricting to image as context. The bold values indicate the best performance obtained for the corresponding dataset among all settings}
\begin{multicols}{2}
\begin{center}
\begin{tabular}{|L{3cm}|C{1cm}|C{1cm}|}
\hline
\multicolumn{3}{|c|}{\textbf{MIL with Negative Bag Margin}}              \\ \hline
Proposals           &  GT  & MCG  \\ \hline
\multicolumn{3}{|c|}{\cellcolor[gray]{0.8}Google RefExp - Val}              \\ \hline
Noisy-Or    & \bf{68.4} & {49.5} \\ \hline
Max          & 66.5 & 48.6 \\ \hline
Image context only         & 65.9 & 48.1 \\ \hline
\multicolumn{3}{|c|}{\cellcolor[gray]{0.8}UNC RefExp - Val}                \\ \hline
Noisy-Or      & \bf{76.9} & {57.3} \\ \hline
Max          & 75.5 & 56.5 \\ \hline
Image context only        & 76.4 & 56.7 \\ \hline
\multicolumn{3}{|c|}{\cellcolor[gray]{0.8}UNC RefExp - TestA}               \\ \hline
Noisy-Or      & 75.6 & 58.6 \\ \hline
Max          & 74.1 & 57.9 \\ \hline
Image context only         & \bf{76.2} & {58.8} \\ \hline
\multicolumn{3}{|c|}{\cellcolor[gray]{0.8}UNC RefExp - TestB}               \\ \hline
Noisy-Or      & \bf{78.0} & \bf{56.4} \\ \hline
Max          & 76.8 & 55.3 \\ \hline
Image context only         & 77.0 & 55.0 \\ \hline
\end{tabular}
\end{center}
\columnbreak
\begin{center}
\begin{tabular}{|L{3.5cm}|C{1cm}|C{1cm}|}
\hline
\multicolumn{3}{|c|}{\textbf{MIL with Pos. \& Neg. Bag Margin}}          \\ \hline
Proposals           &  GT  & MCG  \\ \hline
\multicolumn{3}{|c|}{\cellcolor[gray]{0.8}Google RefExp - Val}              \\ \hline
Noisy-Or    & \bf{68.4} & \bf{50.0} \\ \hline
Max          & 67.2 & 49.3 \\ \hline
Image context only         & 67.9 & 49.3 \\ \hline
\multicolumn{3}{|c|}{\cellcolor[gray]{0.8}UNC RefExp - Val}                \\ \hline
Noisy-Or      & {76.1} & \bf{57.4} \\ \hline
Max          & 75.3 & 56.5 \\ \hline
Image context only         & 76.1 & 56.6 \\ \hline
\multicolumn{3}{|c|}{\cellcolor[gray]{0.8}UNC RefExp - TestA}               \\ \hline
Noisy-Or      & 75.0 & 58.7 \\ \hline
Max          & 73.4 & 58.2 \\ \hline
Image context only         & {75.5} & \bf{58.9} \\ \hline
\multicolumn{3}{|c|}{\cellcolor[gray]{0.8}UNC RefExp - TestB}               \\ \hline
Noisy-Or      & {77.5} & {56.3} \\ \hline
Max          & 76.1 & 55.3 \\ \hline
Image context only         & 76.1 & 55.0 \\ \hline
\end{tabular}
\end{center}
\end{multicols}
\label{Tab:AblationResults}
\end{table}

\subsection{Ablation experiments}
In Table \ref{Tab:SOTAComparison}, the results for the MIL based methods use the noisy-or function for measuring the probability of a referring expression for a region.
The noisy-or function integrates context information from multiple pairs of a regions.
We can also use the max function to determine the probability of a referring expression for a region. 
In this case, the probability for a region is defined as the maximum probability obtained by any of its pairings with other regions.
We also experiment with restricting the context region set to include only the image during testing.

The results in Table \ref{Tab:AblationResults} show that noisy-or pooling provides the best performance on all partitions except the UNC RefExp TestA partition.
It is also more robust when compared to max pooling, which does not exhibit consistent performance.
Our models with just image context perform better than the max-margin model of Mao et al. \cite{Mao15} which also used only image as context.
The reason for this improvement is that our MIL based loss functions mine negative samples for context during training.
In the max-margin model of Mao et al. \cite{Mao15}, the model was trained on negative samples for only the referred region and it was not possible to sample negatives for context.

\begin{figure}[t!]
\centering
\includegraphics[width=\textwidth]{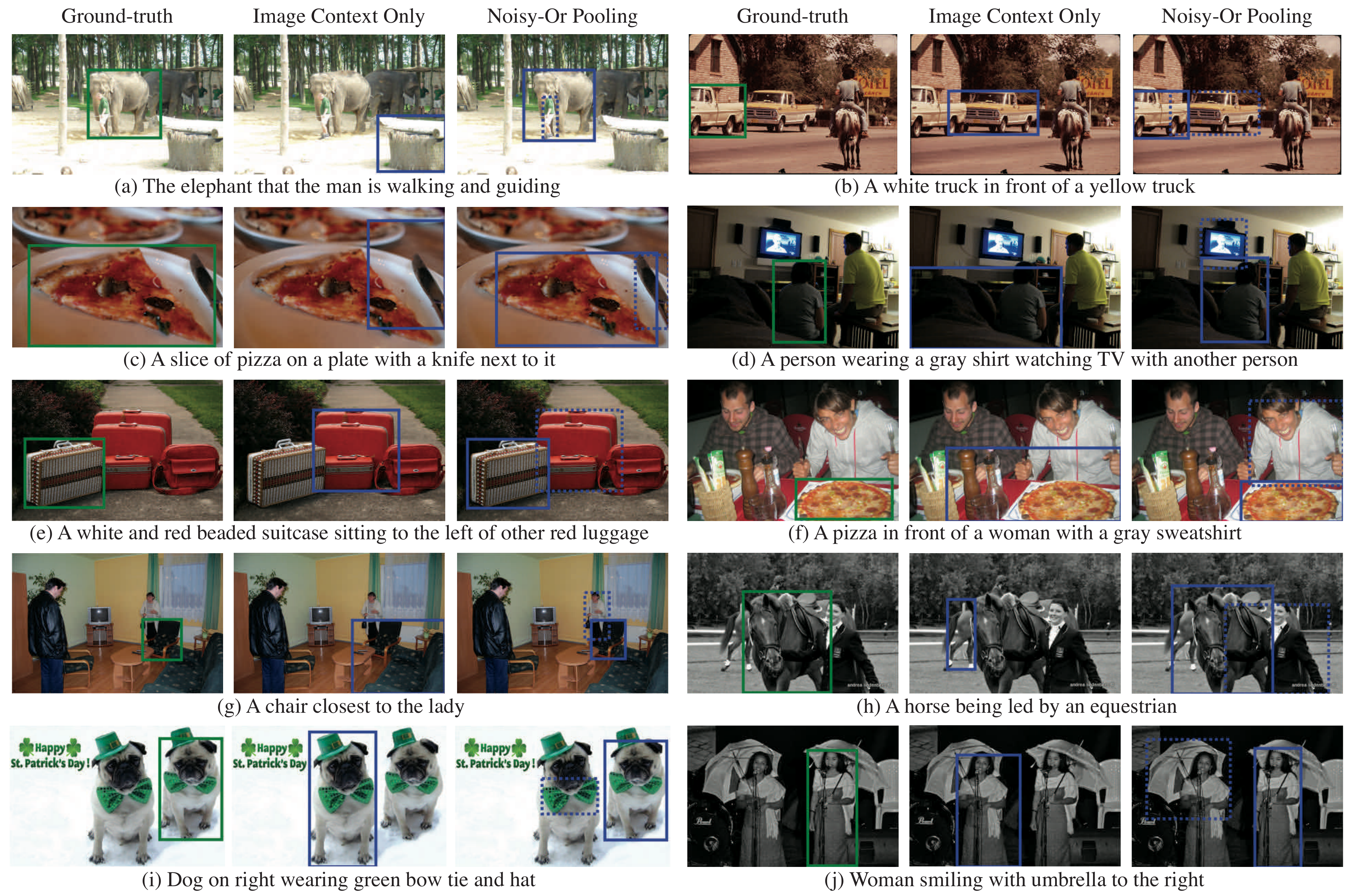}
\caption{Google RefExp results. We show results from the model trained with positive and negative bag margin. We compare the grounding between using image context only and pooling the context from all regions using noisy-or. A box with dashed line indicates the context region. We first identify the referred region using noisy-or function. The context region is then selected as the one which produces maximum probability with the referred region. The last row shows images with misplaced context regions}
\label{Fig:Google_pos_results}
\end{figure}

\begin{figure}[t!]
\centering
\includegraphics[width=\textwidth]{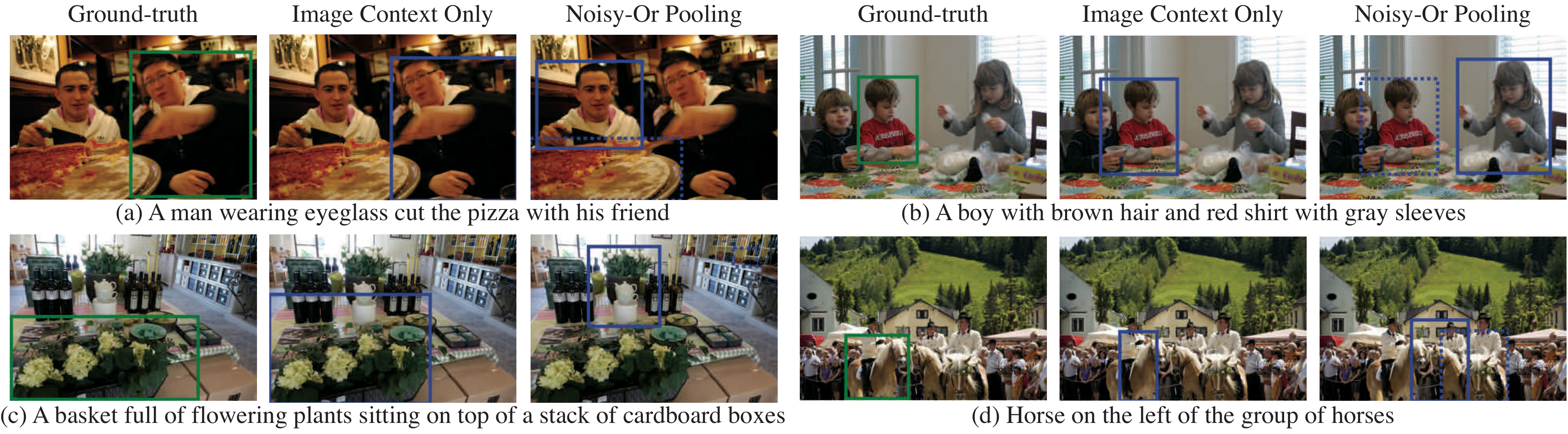}
\caption{Google RefExp failure cases.
We observe errors when there is wrong grounding of attributes or when there is incorrect localization of context region
}
\label{Fig:Google_neg_results}
\end{figure}

Figure \ref{Fig:Google_pos_results} and Figure \ref{Fig:Google_neg_results} show a few sample results from the Google RefExp dataset. 
We observe that our model can localize the referred region and its supporting context region. 
When there is only one instance of an object in an image, the presence of a supporting context region helps in localizing the instance more accurately when compared to using just the image as context.
When there are multiple instances of an object type, the supporting context region resolves ambiguity and helps in localizing the correct instance. 

\begin{figure}[t!]
\centering
\includegraphics[width=\textwidth]{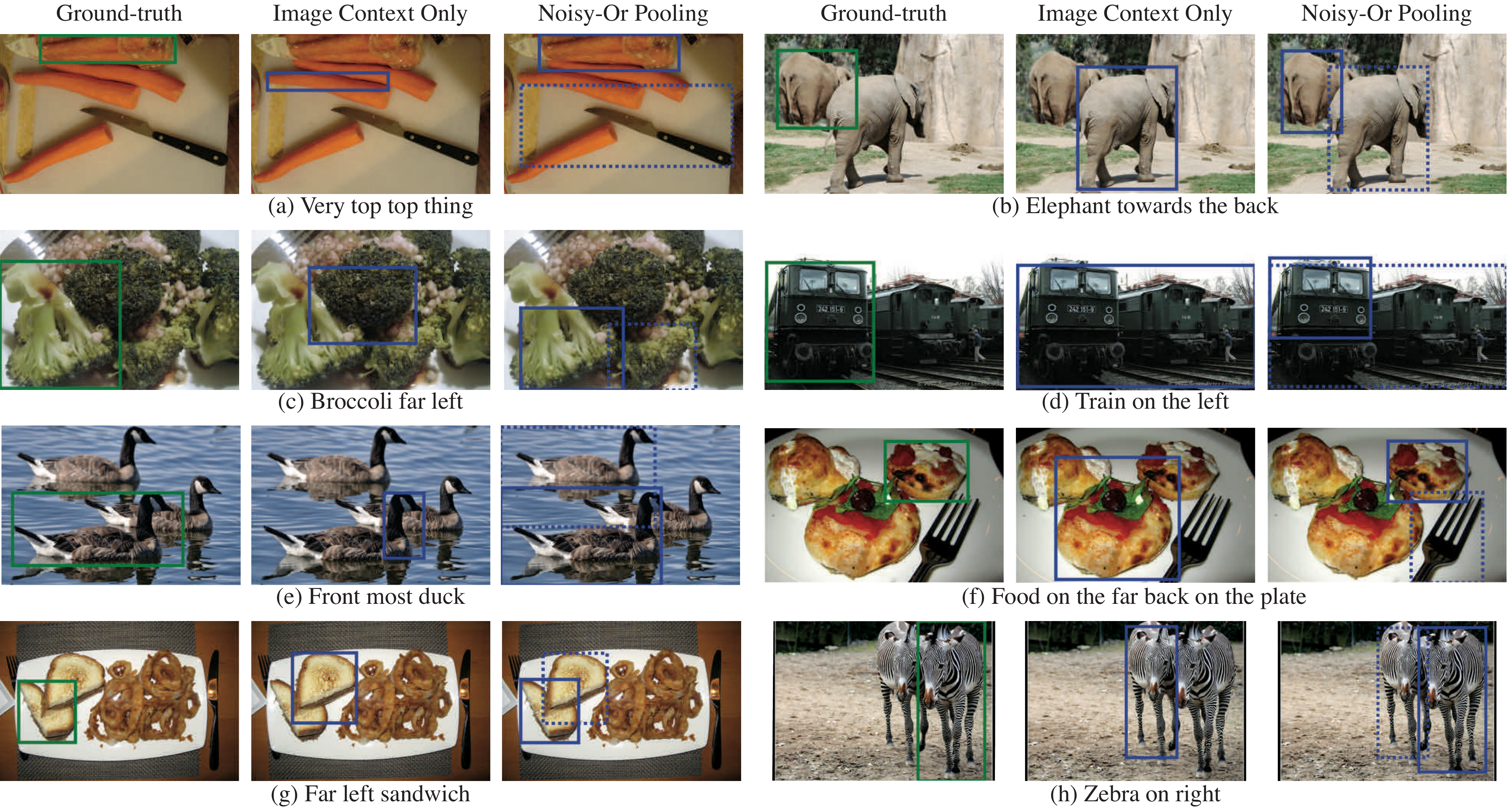}
\caption{UNC RefExp results from TestB partition. We show results from the model trained with negative bag margin. We observe that our method can identify the referred region even when the context object is not explicitly mentioned}
\label{Fig:Berg_pos_results}
\end{figure}

\begin{figure}[t!]
\centering
\includegraphics[width=\textwidth]{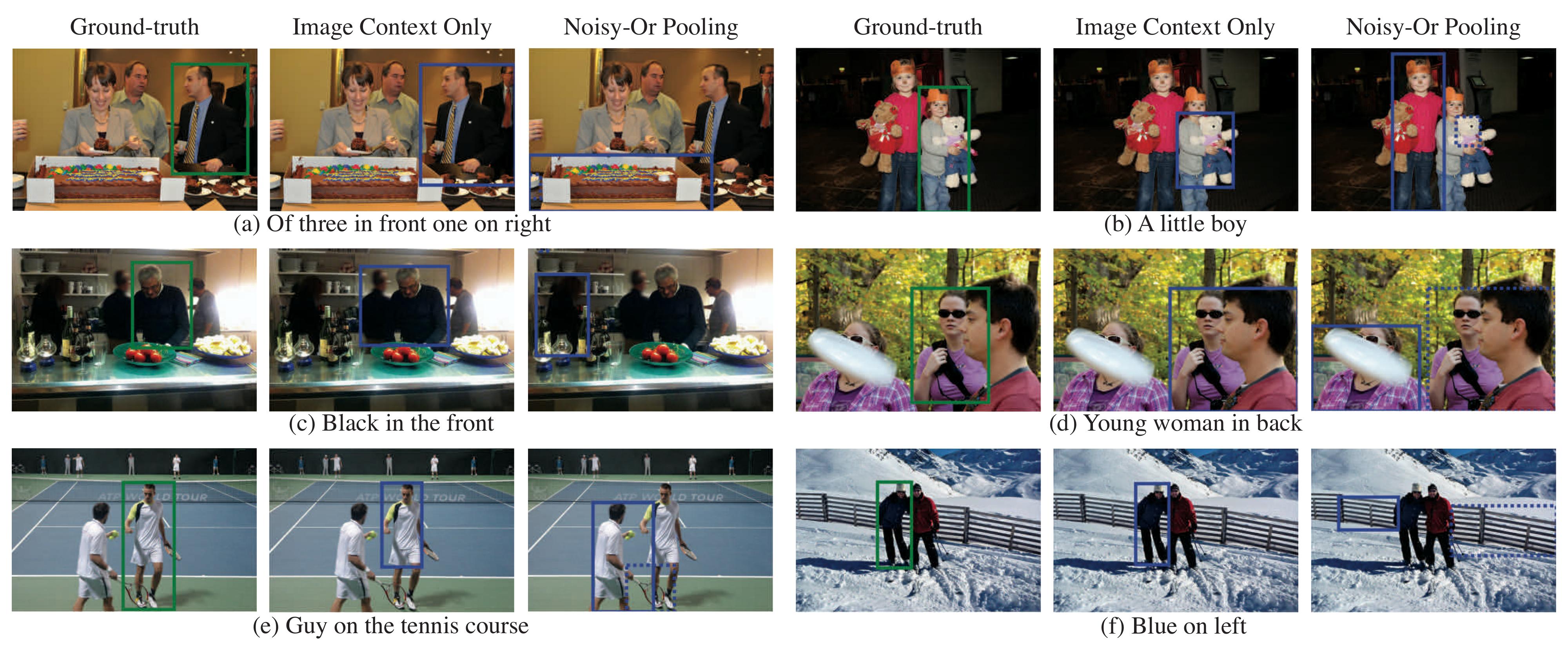}
\caption{UNC RefExp failure cases from TestA partition. We show results from the model trained with negative bag margin. This partition contains terse referring expressions. Most of the time, the referring expressions do not uniquely identify the people}
\label{Fig:Berg_neg_results}
\end{figure}

The sample results in Figure \ref{Fig:Berg_pos_results} from the TestB partition of the UNC RefExp dataset shows that our method can identify the referred region even when the context object is not explicitly mentioned.
Since our method considers pairs of regions, it can evaluate the likelihood of a region relative to another region.
For example, when there are two instance of the same object on the left, our method can evaluate which of those two instances is more to the left than the other.
On the TestA partition of UNC RefExp dataset, we observe that adding context did not improve performance.
Samples from this partition are shown in Figure \ref{Fig:Berg_neg_results}.
The referring expressions in this partition deal with people only and are usually terse.
They do not always refer to a unique region in the image.
We also observe that many referring expressions do not mention that they are referring to a person.

\begin{figure}[t!]
\centering
\includegraphics[width=\textwidth]{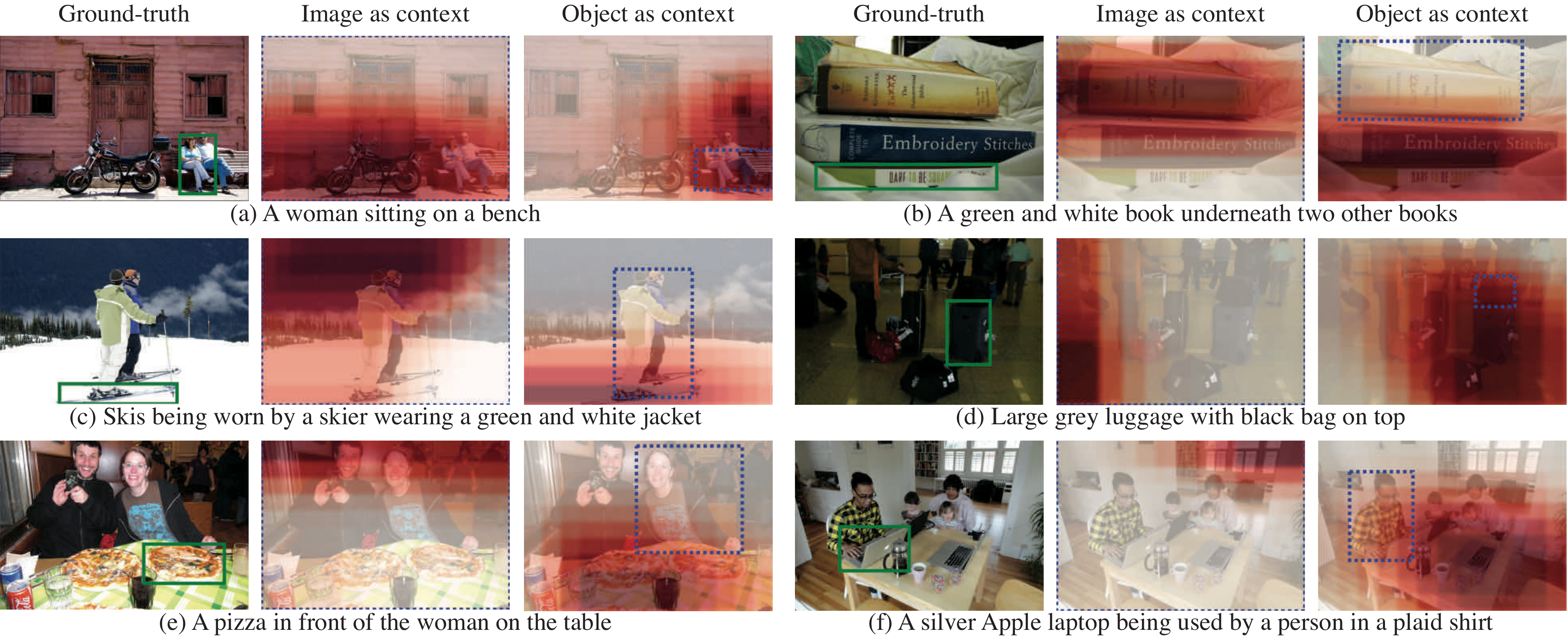}
\caption{Spatial likelihood of referred region given a context region. We fix the context region and evaluate the likelihood of the referred object being present in various locations of the image. When the entire image is used as context, the high likelihood regions do not necessarily overlap with the location of the referred region. However when the context region is fixed, the high likelihood regions overlap with the referred region}
\label{Fig:RelationshipVis}
\end{figure}

To observe the effect of spatial relationships between objects, we move the referred region to different locations in the image and evaluate the likelihood of the referred region at different locations.
Figure \ref{Fig:RelationshipVis} shows sample heat-maps of the likelihood of a referred object.
We first select the entire image as context and observe that the likelihood map is not indicative of the location of the referred object.
However, when the relevant context object is selected, the regions of high likelihood overlap with the location of referred object.

\section{Conclusions}
We have proposed a technique that models the probability of a referring expression as a function of a region and a context region using an LSTM.
We demonstrated that multiple-instance learning based objective functions can be used for training LSTMs to handle the lack of annotations for context objects. 
Our two formulations of the training objective functions are conceptually similar to MISVM and mi-SVM \cite{MISVM}.
The results on Google RefExp and UNC RefExp dataset show that our technique performs better than the max-margin model of Mao et al. \cite{Mao15}.
The qualitative results show that our models can identify a referred region along with its supporting context region.\\

\noindent \textbf{Acknowledgement} This research was supported by contract N00014-13-C-0164 from the Office of Naval Research through a subcontract from the United Technologies Research Center. The GPUs used in this research were donated by the NVIDIA Corporation. We thank Junhua Mao, Licheng Yu and Tamara Berg for helping with the datasets. We also thank Bharat Singh for helpful discussions.

\newpage
\bibliographystyle{splncs}
\bibliography{refexp}
\end{document}